\documentclass[11pt,a4paper]{article}
\usepackage[hyperref]{acl2020}
\usepackage{times}
\usepackage{latexsym}

\usepackage{amsmath}
\usepackage{graphicx}
\usepackage{color}
\usepackage[font=small,labelfont=bf]{caption}
\usepackage{placeins}
\usepackage{wrapfig}
\usepackage{sidecap}
\usepackage{booktabs}
\usepackage{sidecap}
\usepackage{wrapfig}
\usepackage{amssymb}
\usepackage{microtype}
\usepackage{cleveref}

\crefformat{section}{\S#2#1#3} 
\crefformat{subsection}{\S#2#1#3}
\crefformat{subsubsection}{\S#2#1#3}
\aclfinalcopy 

\title{A Probabilistic Generative Model for \\ Typographical Analysis of Early Modern Printing}

\author{ Kartik Goyal$^1$\ \ Chris Dyer$^2$\ \  Christopher Warren$^3$\ \  Max G'Sell$^4$\ \  Taylor Berg-Kirkpatrick$^5$ \\
        $^1$Language Technologies Institute, Carnegie Mellon University\ \ \ $^2$Deepmind \\
        $^3$Department of English, Carnegie Mellon University \\
       $^4$Department of Statistics, Carnegie Mellon University \\
       $^5$Computer Science and Engineering, University of California, San Diego \\
    \texttt{\small \{kartikgo,cnwarren,mgsell\}@andrew.cmu.edu}\ \ \texttt{\small cdyer@google.com}\ \ \texttt{\small tberg@eng.ucsd.edu}}
\date{}

\begin{document}
\maketitle
\begin{abstract}
We propose a deep and interpretable probabilistic generative model to analyze glyph shapes in printed Early Modern documents. We focus on clustering extracted glyph images into underlying templates in the presence of multiple confounding sources of variance.
Our approach introduces a neural editor model that first generates well-understood printing phenomena like spatial perturbations from template parameters via interpertable latent variables, and then modifies the result by generating a non-interpretable latent vector responsible for inking variations, jitter, noise from the archiving process, and other unforeseen phenomena associated with Early Modern printing. Critically, by introducing an inference network whose input is restricted to the visual residual between the observation and the interpretably-modified template, we are able to control and isolate what the vector-valued latent variable captures.
%
We show that our approach outperforms rigid interpretable clustering baselines (Ocular) and overly-flexible deep generative models (VAE) alike on the task of \emph{completely unsupervised} discovery of typefaces in mixed-font documents.  
\end{abstract}

\section{Introduction}
Scholars interested in understanding details related to production and provenance of historical documents rely on methods of analysis ranging from the study of orthographic differences and stylometrics, to visual analysis of layout, font, and printed characters. Recently developed tools like Ocular \cite{berg-kirkpatrick-etal-2013-unsupervised} for OCR of historical documents have helped automate and scale some textual analysis methods for tasks like compositor attribution \cite{ryskina-etal-2017-automatic} and digitization of historical documents \cite{garrette-etal-2015-unsupervised}. However, 
researchers often find the need to go beyond textual analysis for establishing provenance of historical documents. For example, \citet{hinman1963printing}'s study of typesetting in Shakespeare's First Folio relied on the discovery of pieces of damaged or distinctive type through manual inspection of every glyph in the document. More recently, \citet{pnp2020} examine pieces of distinctive types across several printers of the early modern period to posit the identity of clandestine printers of John Milton's \emph{Areopagitica} (1644). In such work, researchers frequently aim to determine whether a book was produced by a single or multiple printers (\citet{weiss_shared_1992,malcolm_editorial_2014,takano_thomas_2016}).  Hence, in order to aid these visual methods of analyses, we propose here a novel probabilistic generative model for analyzing extracted images of individual printed characters in historical documents. We draw from work on both deep generative modeling and interpretable models of the printing press to develop an approach that is both flexible and controllable -- the later being a critical requirement for such analysis tools.

\begin{figure}[t]
\centering
\includegraphics[height=4.0cm]{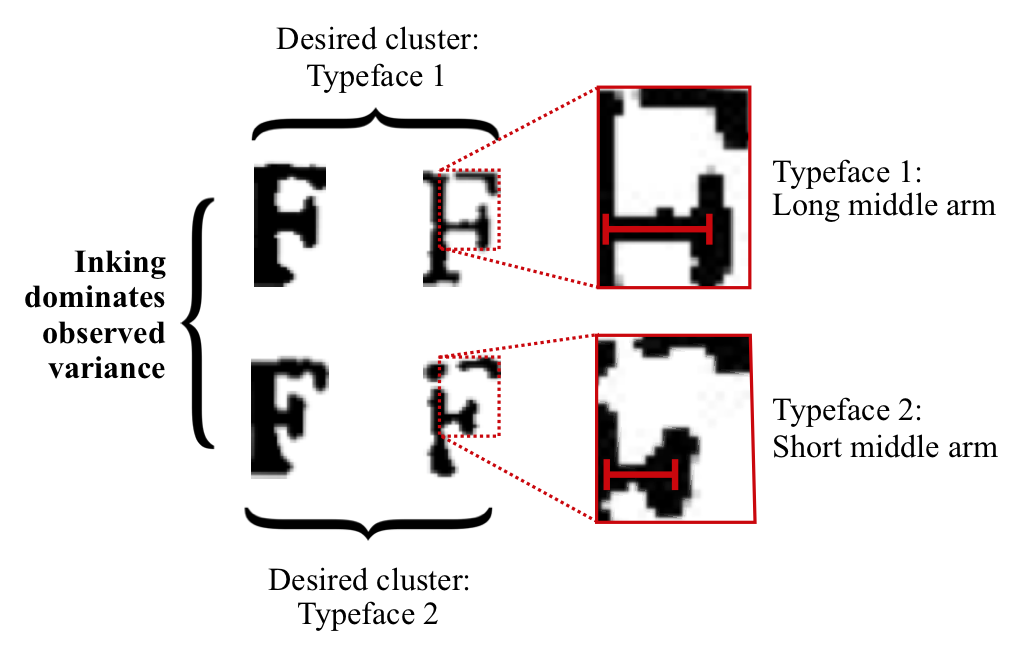}
\vspace{-0.2cm}
  \caption{\label{intro}We desire a generative model that can be biased to cluster according to typeface characteristics (e.g. the length of the middle arm) rather than other more visually salient sources of variation like inking.}
  \vspace{-0.3cm}
\end{figure}
As depicted in Figure~\ref{intro}, we are interested in identifying clusters of subtly distinctive glyph shapes as these correspond to distinct metal stamps in the type-cases used by printers.  However, other sources of variation (inking, for example, as depicted in Figure~\ref{intro}) are likely to dominate conventional clustering methods. For example, powerful models like the variational autoencoder (VAE) \cite{KingmaW13} capture the more visually salient variance in inking rather than typeface, while more rigid models (e.g. the emission model of Ocular \cite{berg-kirkpatrick-etal-2013-unsupervised}), fail to fit the data. The goal of our approach is to account for these confounding sources of variance, while isolating the variables pertinent to clustering. 

Hence, we propose a generative clustering model that introduces a neural editing process to add \emph{expressivity}, but includes \emph{interpretable} latent variables that model well-understood variance in the printing process: bi-axial translation, shear, and rotation of canonical type shapes. In order to make our model \emph{controllable} and prevent deep latent variables from explaining all variance in the data, we introduce a restricted inference network. By only allowing the inference network to observe the visual residual of the observation after interpretable modifications have been applied, we bias the posterior approximation on the neural editor (and thus the model itself) to capture residual sources of variance in the editor -- for example, inking levels, ink bleeds, and imaging noise.  
This approach is related to recently introduced neural editor models for text generation \cite{guu2018generating}. 

In experiments, we compare our model with rigid interpretable models (Ocular) and powerful generative models (VAE) at the task of unsupervised clustering subtly distinct typeface in scanned images early modern documents sourced from Early English Books Online (EEBO).

\section{Model}
\begin{figure}[t]
\centering
\includegraphics[height=8.0cm]{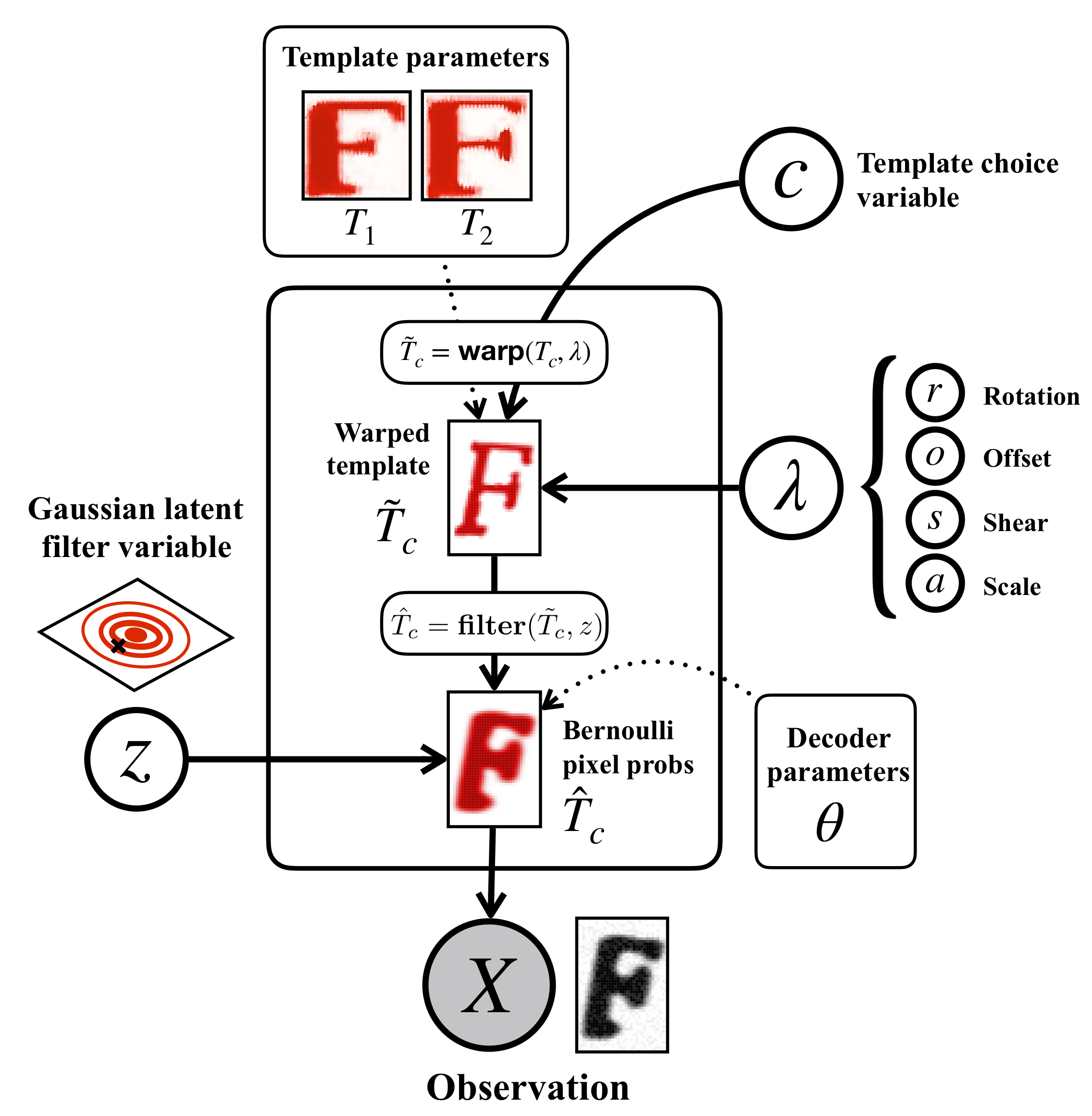}
\vspace{-0.3cm}
 \caption{\label{model}Proposed generative model for clustering images of a symbol by typeface. Each mixture component $c$ corresponds to a \emph{learnable} template $T_k$. The $\lambda$ variables warp (spatially adjust) the original template $T$ to $\tilde{T}$. This warped template is then further transformed via the $z$ variables to $\hat{T}$ via an expressive neural filter function parametrized by $\theta$.
 }
 \vspace{-0.2cm}
\end{figure}

Our model reasons about the printed appearances of a symbol (say majuscule \texttt{F}) in a document via a mixture model whose $K$ components correspond to different metal stamps used by a printer for the document. During various stages of printing,
random transformations result in varying printed manifestations of a metal cast on the paper. Figure~\ref{model} depicts our model. We denote an observed image of the extracted character by $X$.
We denote choice of typeface by latent variable $c$ (the mixture component) with prior $\pi$. We represent the shape of the $k$-th stamp by template $T_k$, a square matrix of parameters. We denote the interpretable latent variables corresponding to spatial adjustment of the metal stamp by $\lambda$, and the editor latent variable responsible for residual sources of variation by $z$. 
As illustrated in Fig.~\ref{model}, after a cluster component $c=k$ is selected, the corresponding template $T_k$ undergoes a transformation to yield $\hat{T_k}$. This transformation occurs in two stages: first, the interpretable spatial adjustment variables ($\lambda$) produce an adjusted template (\cref{lbd}), $\tilde{T_k} = \textrm{warp}(T_k, \lambda)$, and then the neural latent variable transforms the adjusted template (\cref{zvec}), $\hat{T_k} = \textrm{filter}(\tilde{T_k},z)$.
The marginal probability under our model can be written as
\begin{align*}
p(X) = \sum_k \pi_k \int p(X|\lambda, z; T_k)p(\lambda) p(z)  dz d\lambda
\end{align*}
where $p(X|\lambda, z; T_k)$ refers to the distribution over the binary pixels of $X$ where each pixel has a bernoulli distribution parametrized by the value of the corresponding pixel-entry in $\hat{T_k}$. 
\subsection{Interpretable spatial adjustment}
\label{lbd}
Early typesetting was noisy, and the metal pieces 
were often arranged with slight variations which resulted in the printed characters being positioned with small amounts of offset, rotation and shear. 
These real-valued spatial adjustment variables are denoted by $\lambda=(r, o, s, a)$, where $r$ represents the rotation variable, $o = (o_h,o_v)$ represents offsets along the horizontal and vertical axes, $s = (s_h,s_v)$ denotes shear along the two axes. A scale factor, $\tilde{a}= 1.0+a$, accounts for minor scale variations arising due to the archiving and extraction processes. All variables in $\lambda$ are generated from a Gaussian prior with zero mean and fixed variance as the transformations due to these variables tend to be subtle.

\begin{figure}[h]
\centering
\includegraphics[height=2.1cm]{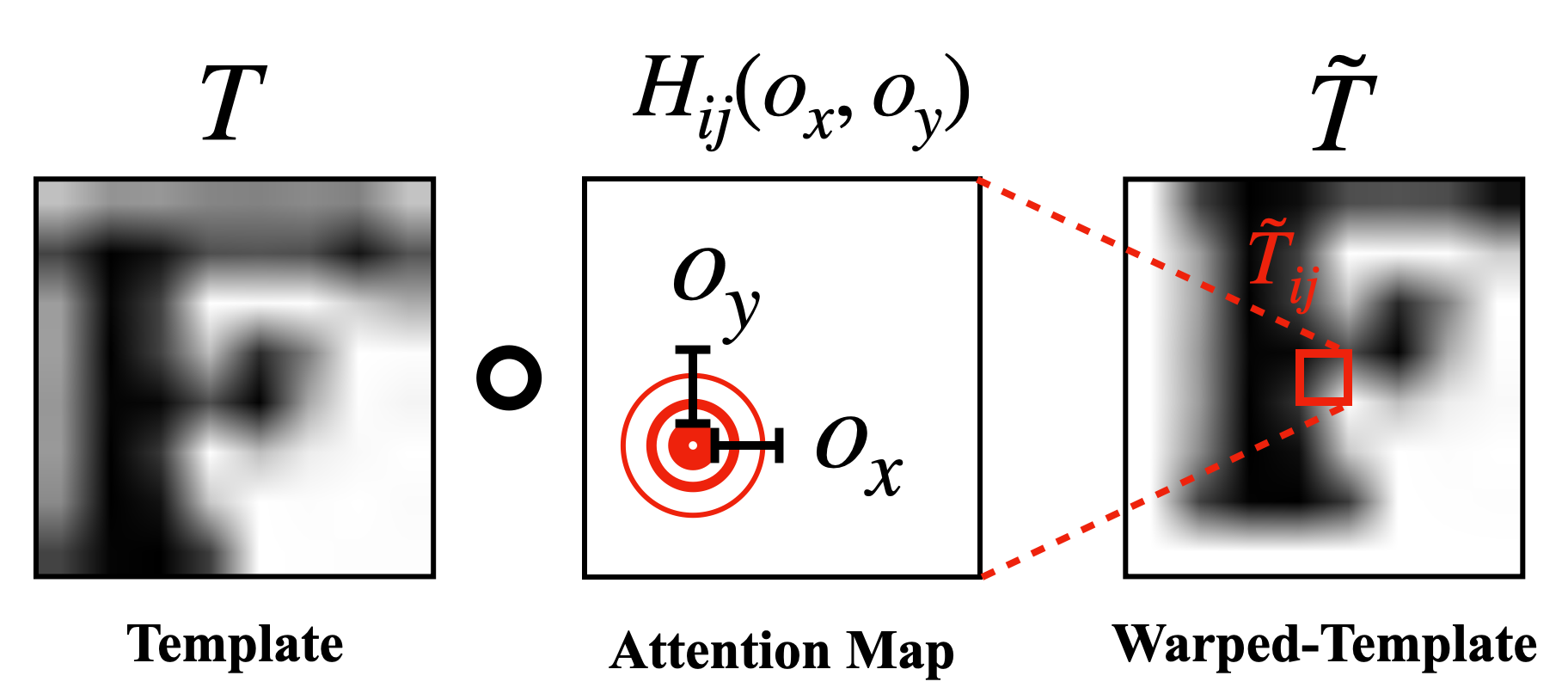}
\vspace{-0.2cm}
  \caption{\label{heatmap} Translation operation: The mode of the attention map is shifted by the offset values for every output pixel in $\tilde{T}$. Similar operations account for shear, rotation, and scale.}
  \vspace{-0.3cm}
\end{figure}

In order to incorporate these deterministic transformations in a differentiable manner, we map $\lambda$ to a template sized attention map $H_{ij}$ for each output pixel position $(i,j)$ in $\tilde{T}$ as depicted in Figure~\ref{heatmap}. The attention map for each output pixel is formed in order to attend to the corresponding shifted (or scaled or sheared) portion of the input template and is shaped according to a Gaussian distribution with mean determined by an affine transform. 
This approach allows for strong inductive bias which contrasts with related work on \emph{spatial-VAE} \cite{bepler2019explicitly} that learns arbitrary transformations.
\subsection{Residual sources of variations}
\label{zvec}
Apart from spatial perturbations, other major sources of deviation in early printing include random inking perturbations caused by inconsistent application of the stamps, unpredictable ink bleeds, and noise associated with digital archiving of the documents. Unlike in the case of spatial perturbations which could be handled by deterministic affine transformation operators, it is not possible to analytically define a transformation operator due to these variables. Hence we propose to introduce a non-interpretable real-valued latent vector $z$, with a Gaussian prior $\mathcal{N}(\mathbf{0},\mathbf{I})$ 
, that transforms $\tilde{T}$ into a final template $\hat{T}$ via neurally-parametrized function $\textrm{filter}(\tilde{T},z; \theta)$ with neural network parameters $\theta$. This function is a convolution over $\tilde{T}$ whose kernel is parametrized by $z$, followed by non-linear operations. Intuitively, parametrizing the filter by $z$ results in the latent variable accounting for variations like inking appropriately because convolution filters capture local variations in appearance. \citet{srivatsan-etal-2019-deep} also observed the effectiveness of using $z$ to define a deconvolutional kernel for font generation.
\subsection{Learning and Inference}
Our aim is to maximize the log likelihood of the observed data ($\{X_d\mid d \in \mathbb{N}, d<n\}$) of $n$ images wrt. model parameters:
%

\begin{align*}
\textrm{LL}(T_{1,\ldots,k}, \theta) =\max_{T,\theta} \sum_{d} \log \Big[\sum_k \pi_k \\ \int p(X_d|\lambda_d, z_d; T_k,\theta)p(\lambda_d) p(z_d)  dz_d d\lambda_d \Big]
\end{align*}
 
\noindent During training, we maximize the likelihood wrt. $\lambda$ instead of marginalizing, which is an approximation inspired by iterated conditional modes \cite{besag1986statistical}:
\begin{align*}
 \max_{T,\theta} \sum_{d} \log \sum_k \max_{\gamma_{k,d}} \pi_k  \int p(X_d|\lambda_d=\gamma_{k,d}, z_d;\\ T_k, \theta) p(\lambda_d=\gamma_{k,d}) p(z_d) dz_d   
\end{align*}
\begin{figure}[t]
\centering
\includegraphics[height=5.4cm]{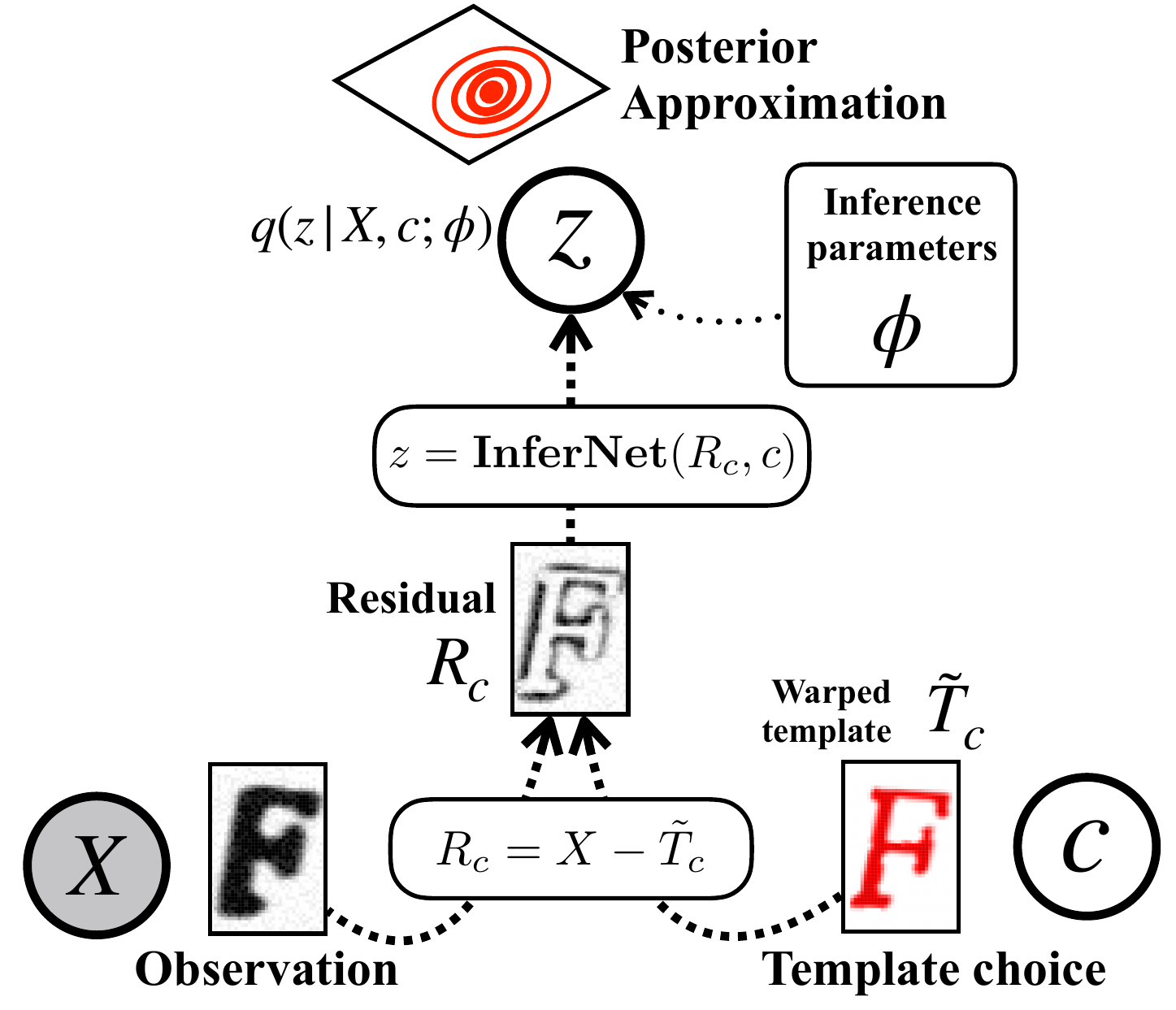}
\vspace{-0.2cm}
  \caption{\label{infnet} Inference network for $z$ conditions on the mixture component and only the residual image left after subtracting the $\lambda$-transformed template from the image. This encourages $z$ to model variance due to sources other than spatial adjustments.}
  \vspace{-0.3cm}
\end{figure}
However, marginalizing over $z$ remains intractable. Therefore we perform amortized variational inference to define and maximize a lower bound on the above objective \cite{KingmaW13}. We use a convolutional inference neural network parametrized by $\phi$ (Fig.~\ref{infnet}), that takes as input, the mixture component $k$, the residual image $R_k = X-\tilde{T_k}$, and produces mean and variance parameters for an isotropic gaussian proposal distribution $q(z \mid R_k,k; \phi )$. This results in the final training objective:
\begin{align*}
 \max_{T,\theta, \phi} \sum_{d} \log \sum_k \mathrm{E}_{q(z_{d}\mid R_{d,k},k;\phi)}\big[ \max_{\gamma_{k,d}} \big(\pi_k \\p(X_d|\lambda = \gamma_{k,d}, z_{d}; T_k, \theta) p(\lambda = \gamma_{k,d})\big)\big]\\ - \textrm{KL}\big(q(z_{d}|R_{d,k},k; \phi) || p(z)\big)
\end{align*}
 
\noindent We use stochastic gradient ascent to maximize this objective with respect to $T, \gamma, \theta$ and $\phi$.

\section{Experiments}
We train our models on printed occurrences of 10 different uppercase character classes that scholars have found useful for bibliographic analysis \cite{pnp2020} because of their distinctiveness.
As a preprocessing step, we ran Ocular \cite{berg-kirkpatrick-etal-2013-unsupervised} on the grayscale scanned images of historical books in EEBO dataset and extracted the estimated image segments for the letters of interest.

\subsection{Quantitative analysis}
We show that our model is superior to strong baselines at clustering subtly distinct typefaces (using realistic synthetic data), as well as in terms of fitting the real data from historical books. 
\subsubsection{Baselines for comparison}
\noindent \textbf{Ocular:} Based on the emission model of Ocular that uses discrete latent variables for the vertical/horizontal offset and inking variables, and hence has limited expressivity.\\
\noindent \textbf{$\lambda$-only:} This model only has the interpretable continuous latent variables pertaining to spatial adjustment.\\
\noindent \textbf{VAE-only:} This model is expressive but doesn't have any interpretable latent variables for explicit control. It is an extension of \citet{kingma2014semi}'s model for semi-supervised learning with a continuous latent variable vector in which we obtain tighter bounds by marginalizing over the cluster identities explicitly. 
For fair comparison, the encoder and decoder convolutional architectures are the same as the ones in our full model. The corresponding training objective for this baseline can be written as:
\begin{align*}
 \max_{T,\theta, \phi} \sum_{d} \log \sum_k \mathrm{E}_{q(z_{d}\mid X_{d},k;\phi)}\big[\pi_k p(X_d| z_{d}; T_k, \theta)\big]\\ - \textrm{KL}\big(q(z_{d}|X_{d},k; \phi) || p(z)\big)
\end{align*}
\noindent \textbf{No-residual:} The only difference from the full model is that the encoder for the inference network conditions the variational distribution $q(z)$ on the entire input image $X$ instead of just the residual image $X - \tilde{T}$.
\subsubsection{Font discovery in Synthetic Data}
Early modern books were frequently composed from two or more type cases,
resulting in documents with mixed fonts. We aim to learn the different shapes of metal stamps that were used as templates for each cluster component in our model.\\
\noindent \textbf{Data:} In order to quantitatively evaluate our model's performance, we experiment with synthetically generated realistic dataset for which we know the ground truth cluster identities in the following manner: For each character of interest, we pick three distinct images from scanned segmented EEBO images, corresponding to three different metal casts. Then we randomly add spatial perurbations related to scale, offset, rotation and shear. To incorporate varying inking levels and other distortions, we randomly either perform \emph{erosion}, \emph{dilation}, or a combination of these warpings using OpenCV \cite{opencv_library} with randomly selected kernel sizes. Finally, we add a small Gaussian noise to the pixel intensities and generate 300 perturbed examples per character class.\\
\noindent \textbf{Results:}
We report macro-averaged results across all the character classes on three different clustering measures, $\textrm{V-measure}$ \cite{rosenberg2007v}, $\textrm{Mutual Information}$ and $\textrm{Fowlkes and Mallows Index}$ \cite{fowlkes1983method}. In Table~\ref{results}, we see that our model significantly outperforms all other baselines on every metric. \emph{Ocular} and \emph{$\lambda$-only} models fail because they lack expressiveness to explain the variations due to random jitters, erosions and dilations. The \emph{VAE-only} model, while very expressive, performs poorly because it lacks the inductive bias needed for successful clustering. The \emph{No-residual} model performs decently but our model's superior performance emphasizes the importance of designing a restrictive inference network such that $z$ doesn't explain any variation due to the interpretable variables.
\hspace{-0.2cm}\begin{table}[t]
\centering
\scalebox{0.8}{
\begin{tabular}{llll|l}
\toprule
 & \textbf{V-measure}          & \textbf{Mutual Info}        & \textbf{F\&M} & \textbf{NLL}         \\
\midrule
\textbf{Ocular} & 0.42 & 0.45 & 0.61 & 379.21\\
\textbf{$\lambda$-only} & 0.49 & 0.51 & 0.70 & 322.04\\
\textbf{VAE-only} & 0.22 & 0.29 & 0.38 & 263.45\\
\textbf{No-residual} & 0.54 & 0.58 & 0.73 & 264.27\\
\textbf{Our Model} & \textbf{0.73} & \textbf{0.74} & \textbf{0.85} & \textbf{257.92}\\
\bottomrule

\end{tabular}
}
\caption{\label{results} (a) Clustering results on synthetic data (\textbf{V-measure}, \textbf{Mutual Info}, \textbf{F\&M}). (b) Test negative log likelihood (NLL) on real data from historical documents, or negative ELBO bound for intractable models (\textbf{NLL}).}
\vspace{-0.2in}
\end{table}

\subsubsection{Fitting Real Data from Historical Books}
For the analysis of real books, we selected three 
books from the EEBO dataset printed by different printers.
We modeled each character class for each book separately and report the macro-aggregated upper bounds on the negative log likelihood (NLL) in Table~\ref{results}. We observe that adding a small amount of expressiveness makes our \emph{$\lambda$-only} model better than \emph{Ocular}. The upper bounds of other inference network based models are much better than the (likely tight)\footnote{For \textbf{Ocular} and \textbf{$\lambda$-only} models, we report the upper bound obtained via maximization over the interpretable latent variables. Intuitively, these latent variables are likely to have unimodal posterior distributions with low variance, hence this approximation is likely tight.} bounds of both the interpretable models. Our model has the lowest upper bound of all the models while retaining interpretability and control.

\subsection{Qualitative analysis}
We provide visual evidence of desirable behavior of our model on collections of character extractions from historical books with mixed fonts. Specifically, we discus the performance of our model on the mysterious edition of Thomas Hobbes' \emph{Leviathan} known as ``the 25 Ornaments" edition. \cite{hobbes_leviathan_1651}.  The 25 Ornaments \emph{Leviathan} is an interesting test case for several reasons.  While its title page indicates a publisher and year of publication, both are fabricated \cite{malcolm_editorial_2014}.  The identities of its printer(s) remain speculative, and the actual year of publication is uncertain. Further, the 25 Ornaments exhibits two distinct fonts.  
\subsubsection{Quality of learned templates}
\begin{figure}[h]
\centering
\includegraphics[height=2.9cm]{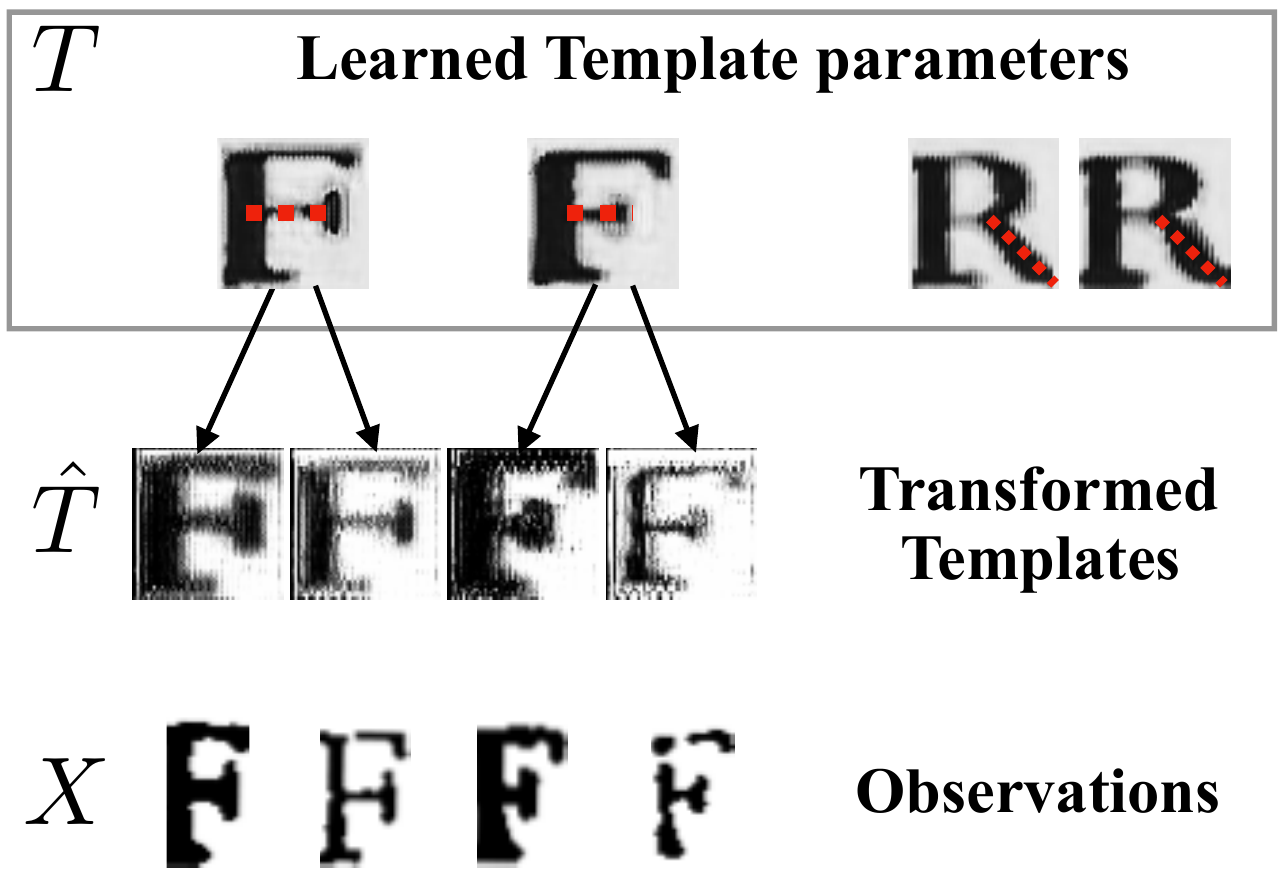}
\vspace{-0.2cm}
  \caption{\label{templates} The learned templates for \textbf{F} and \textbf{R} and the transformed templates $\hat{T}$ for four examples of \textbf{F} are shown. Our model is able to learn desirable templates based on underlying glyph structure.}
  \vspace{-0.2cm}
\end{figure}
Our model is successful in discovering distinctly shaped typefaces in the 25 Ornaments \emph{Leviathan}. We focus on the case study of majuscule letters \textbf{F} and \textbf{R}, each of which have two different typefaces mixed in throughout. The two typefaces for \textbf{F} differ in the \emph{length of the middle arm} (Fig.~\ref{intro}), and the two typefaces for \textbf{R} have \emph{differently shaped legs}. In Fig.~\ref{templates}, we show that our model successfully learns the two desired templates $T_1$ and $T_2$ for both the characters which indicates that the clusters in our model mainly focus on subtle differences in underlying glyph shapes. 
We also illustrate how the latent variables transform the model templates $T$ to $\hat{T}$ for four example \textbf{F} images. The model learns complex functions to transform the templates which go beyond simple affine and morphological transformations in order to account for inking differences, random jitter, contrast variations etc. 
\subsubsection{Interpretable variables ($\lambda$) and Control}
\begin{figure}[h]
\centering
\includegraphics[height=1.95cm]{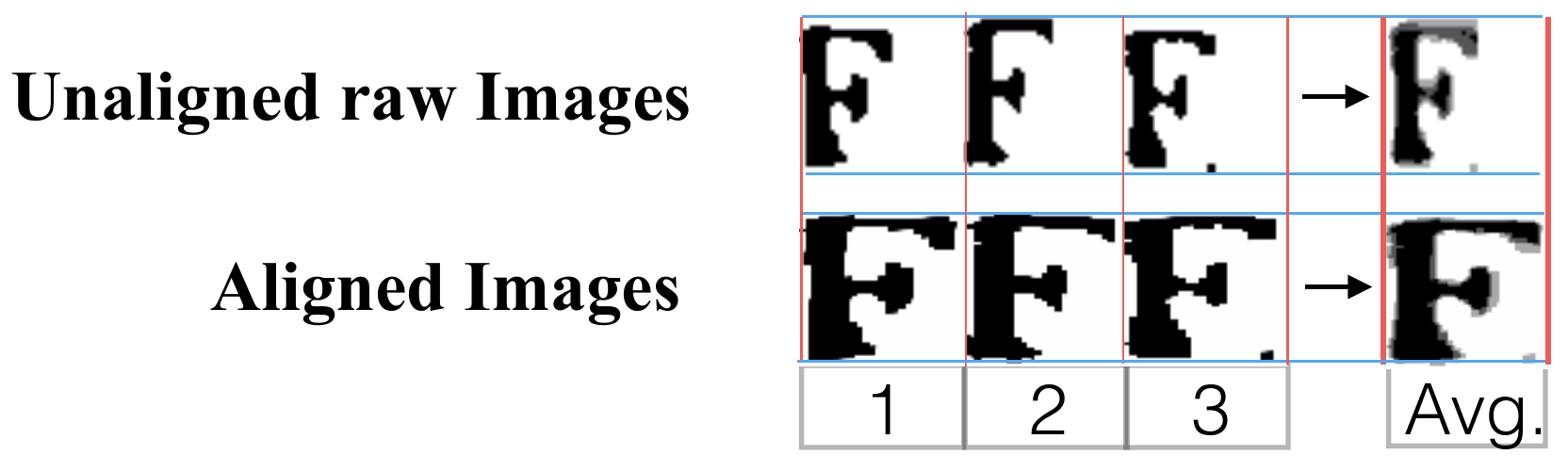}
\vspace{-0.2cm}
  \caption{\label{alignment} Result of alignment on \emph{Leviathan} extractions using the interpretable $\lambda$ variables along with their pixelwise average images. Aligned average image is much sharper than the unaligned average image.}
  \vspace{-0.2cm}
\end{figure}
Finally, we visualize the ability of our model to separate responsibility of modelling variation among the interpretable and non-interpretable variables appropriately. We use the inferred values of the interpretable ($\lambda$) variable for each image in the dataset to adjust the corresponding image. Since the templates represent the canonical shape of the letters, the $\lambda$ variables which shift the templates to explain the images can be \emph{reverse applied} to the input images themselves in order to \emph{align} them by accounting for offset, rotation, shear and minor size variations. In Fig.~\ref{alignment}, we see that the input images (top row) are uneven and vary by size and orientation. By reverse applying the inferred $\lambda$ values, we are able to project the images to a fixed size such that they are aligned and any remaining variations in the data are caused by other sources of variation. Moreover, this \emph{alignment} method would be crucial for automating certain aspects of bibliographic studies that focus on comparing specific imprints.
\section{Conclusion}
Beyond applications to typeface clustering, the general approach we take might apply more broadly to other clustering problems, and the model we developed might be incorporated into OCR models for historical text.
\section{Acknowledgements}

This project is funded in part by the NSF under
grants 1618044 and 1936155, and by the NEH under grant HAA256044-17.

\bibliography{anthology,acl2020}

\begin{thebibliography}{18}
\expandafter\ifx\csname natexlab\endcsname\relax\def\natexlab#1{#1}\fi

\bibitem[{Bepler et~al.(2019)Bepler, Zhong, Kelley, Brignole, and
  Berger}]{bepler2019explicitly}
Tristan Bepler, Ellen Zhong, Kotaro Kelley, Edward Brignole, and Bonnie Berger.
  2019.
\newblock Explicitly disentangling image content from translation and rotation
  with spatial-vae.
\newblock In \emph{Advances in Neural Information Processing Systems}, pages
  15409--15419.

\bibitem[{Berg-Kirkpatrick et~al.(2013)Berg-Kirkpatrick, Durrett, and
  Klein}]{berg-kirkpatrick-etal-2013-unsupervised}
Taylor Berg-Kirkpatrick, Greg Durrett, and Dan Klein. 2013.
\newblock \href {https://www.aclweb.org/anthology/P13-1021} {Unsupervised
  transcription of historical documents}.
\newblock In \emph{Proceedings of the 51st Annual Meeting of the Association
  for Computational Linguistics (Volume 1: Long Papers)}, pages 207--217,
  Sofia, Bulgaria. Association for Computational Linguistics.

\bibitem[{Besag(1986)}]{besag1986statistical}
Julian Besag. 1986.
\newblock On the statistical analysis of dirty pictures.
\newblock \emph{Journal of the Royal Statistical Society: Series B
  (Methodological)}, 48(3):259--279.

\bibitem[{Bradski(2000)}]{opencv_library}
G.~Bradski. 2000.
\newblock {The OpenCV Library}.
\newblock \emph{Dr. Dobb's Journal of Software Tools}.

\bibitem[{Fowlkes and Mallows(1983)}]{fowlkes1983method}
Edward~B Fowlkes and Colin~L Mallows. 1983.
\newblock A method for comparing two hierarchical clusterings.
\newblock \emph{Journal of the American statistical association},
  78(383):553--569.

\bibitem[{Garrette et~al.(2015)Garrette, Alpert-Abrams, Berg-Kirkpatrick, and
  Klein}]{garrette-etal-2015-unsupervised}
Dan Garrette, Hannah Alpert-Abrams, Taylor Berg-Kirkpatrick, and Dan Klein.
  2015.
\newblock \href {https://doi.org/10.3115/v1/N15-1109} {Unsupervised
  code-switching for multilingual historical document transcription}.
\newblock In \emph{Proceedings of the 2015 Conference of the North {A}merican
  Chapter of the Association for Computational Linguistics: Human Language
  Technologies}, pages 1036--1041, Denver, Colorado. Association for
  Computational Linguistics.

\bibitem[{Guu et~al.(2018)Guu, Hashimoto, Oren, and Liang}]{guu2018generating}
Kelvin Guu, Tatsunori~B Hashimoto, Yonatan Oren, and Percy Liang. 2018.
\newblock Generating sentences by editing prototypes.
\newblock \emph{Transactions of the Association for Computational Linguistics},
  6:437--450.

\bibitem[{Hinman(1963)}]{hinman1963printing}
Charlton Hinman. 1963.
\newblock \emph{The printing and proof-reading of the first folio of
  Shakespeare}, volume~1.
\newblock Oxford: Clarendon Press.

\bibitem[{Hobbes(1651 [really 1700?])}]{hobbes_leviathan_1651}
Thomas Hobbes. 1651 [really 1700?].
\newblock \emph{Leviathan, or, the matter, form, and power of a common-wealth
  ecclesiastical and civil. {By} {Thomas} {Hobbes} of {Malmesbury}}.
\newblock Number R13935 in {ESTC}. [false imprint] printed for Andrew Crooke,
  at the Green Dragon in St. Pauls Church-yard, London.

\bibitem[{Kingma and Welling(2014)}]{KingmaW13}
Diederik~P. Kingma and Max Welling. 2014.
\newblock \href {http://arxiv.org/abs/1312.6114} {Auto-encoding variational
  bayes}.
\newblock In \emph{2nd International Conference on Learning Representations,
  {ICLR} 2014, Banff, AB, Canada, April 14-16, 2014, Conference Track
  Proceedings}.

\bibitem[{Kingma et~al.(2014)Kingma, Mohamed, Rezende, and
  Welling}]{kingma2014semi}
Durk~P Kingma, Shakir Mohamed, Danilo~Jimenez Rezende, and Max Welling. 2014.
\newblock Semi-supervised learning with deep generative models.
\newblock In \emph{Advances in neural information processing systems}, pages
  3581--3589.

\bibitem[{Malcolm(2014)}]{malcolm_editorial_2014}
Noel Malcolm. 2014.
\newblock Editorial {Introduction}.
\newblock In \emph{Leviathan}, volume~1. Clarendon Press, Oxford.

\bibitem[{Rosenberg and Hirschberg(2007)}]{rosenberg2007v}
Andrew Rosenberg and Julia Hirschberg. 2007.
\newblock V-measure: A conditional entropy-based external cluster evaluation
  measure.
\newblock In \emph{Proceedings of the 2007 joint conference on empirical
  methods in natural language processing and computational natural language
  learning (EMNLP-CoNLL)}, pages 410--420.

\bibitem[{Ryskina et~al.(2017)Ryskina, Alpert-Abrams, Garrette, and
  Berg-Kirkpatrick}]{ryskina-etal-2017-automatic}
Maria Ryskina, Hannah Alpert-Abrams, Dan Garrette, and Taylor Berg-Kirkpatrick.
  2017.
\newblock \href {https://doi.org/10.18653/v1/P17-2065} {Automatic compositor
  attribution in the first folio of shakespeare}.
\newblock In \emph{Proceedings of the 55th Annual Meeting of the Association
  for Computational Linguistics (Volume 2: Short Papers)}, pages 411--416,
  Vancouver, Canada. Association for Computational Linguistics.

\bibitem[{Srivatsan et~al.(2019)Srivatsan, Barron, Klein, and
  Berg-Kirkpatrick}]{srivatsan-etal-2019-deep}
Nikita Srivatsan, Jonathan Barron, Dan Klein, and Taylor Berg-Kirkpatrick.
  2019.
\newblock \href {https://doi.org/10.18653/v1/D19-1225} {A deep factorization of
  style and structure in fonts}.
\newblock In \emph{Proceedings of the 2019 Conference on Empirical Methods in
  Natural Language Processing and the 9th International Joint Conference on
  Natural Language Processing (EMNLP-IJCNLP)}, pages 2195--2205, Hong Kong,
  China. Association for Computational Linguistics.

\bibitem[{Takano(2016)}]{takano_thomas_2016}
Akira Takano. 2016.
\newblock Thomas {Warren}: {A} {Printer} of {Leviathan} (head edition).
\newblock \emph{Annals of Nagoya University Library Studies}, 13:1--17.

\bibitem[{Warren et~al.(2020)Warren, Williams, Rijhwani, and G'Sell}]{pnp2020}
Christopher~N. Warren, Pierce Williams, Shruti Rijhwani, and Max G'Sell. 2020.
\newblock Damaged type and {A}reopagitica's clandestine printers.
\newblock \emph{Milton Studies}, 62.1.

\bibitem[{Weiss(1992)}]{weiss_shared_1992}
Adrian Weiss. 1992.
\newblock \href {http://www.jstor.org/stable/40371958} {Shared {Printing},
  {Printer}'s {Copy}, and the {Text}(s) of {Gascoigne}'s "{A} {Hundreth}
  {Sundrie} {Flowres}"}.
\newblock \emph{Studies in Bibliography}, 45:71--104.

\end{thebibliography}
\bibliographystyle{acl_natbib}
\appendix

\section{Character wise quantitative analysis}
The quantitative experiments were performed on the following character classes: A, B, E, F, G, H, M, N, R, W.
\begin{table}[h]
\centering
\scalebox{0.8}{
\begin{tabular}{llll|l}
\toprule
 & \textbf{V-measure}          & \textbf{Mutual Info}        & \textbf{F\&M} & \textbf{NLL}         \\
\midrule
\textbf{$\lambda$-only} & 0.77 & 0.82 & 0.89 & 264.90\\
\textbf{VAE-only} & 0.33 & 0.38 & 0.5 & 230.45\\
\textbf{No-residual} & 0.79 & 0.85 & 0.90 & 231.45\\
\textbf{Our Model} & \textbf{0.78} & \textbf{0.86} & \textbf{0.89} & \textbf{226.25}\\
\bottomrule

\end{tabular}
}
\caption{\label{results} Results for character A}
\end{table}

\begin{table}[h]
\centering
\scalebox{0.8}{
\begin{tabular}{llll|l}
\toprule
 & \textbf{V-measure}          & \textbf{Mutual Info}        & \textbf{F\&M} & \textbf{NLL}         \\
\midrule
\textbf{$\lambda$-only} & 0.37 & 0.39 & 0.59 & 261.1\\
\textbf{VAE-only} & 0.15 & 0.2 & 0.32 & 229.1\\
\textbf{No-residual} & 0.37 & 0.39 & 0.58 & 228.1\\
\textbf{Our Model} & \textbf{0.68} & \textbf{0.73} & \textbf{0.81} & \textbf{226.25}\\
\bottomrule

\end{tabular}
}
\caption{\label{results} Results for character B}
\end{table}

\begin{table}[h]
\centering
\scalebox{0.8}{
\begin{tabular}{llll|l}
\toprule
 & \textbf{V-measure}          & \textbf{Mutual Info}        & \textbf{F\&M} & \textbf{NLL}         \\
\midrule
\textbf{$\lambda$-only} & 0.33 & 0.36 & 0.55 & 282.4\\
\textbf{VAE-only} & 0.17 & 0.19 & 0.30 & 253.2\\
\textbf{No-residual} & 0.33 & 0.35 & 0.56 & 251.45\\
\textbf{Our Model} & \textbf{0.65} & \textbf{0.70} & \textbf{0.76} & \textbf{234.05}\\
\bottomrule

\end{tabular}
}
\caption{\label{results} Results for character E}
\end{table}

\begin{table}[h]
\centering
\scalebox{0.8}{
\begin{tabular}{llll|l}
\toprule
 & \textbf{V-measure}          & \textbf{Mutual Info}        & \textbf{F\&M} & \textbf{NLL}         \\
\midrule
\textbf{$\lambda$-only} & 0.09 & 0.10 & 0.55 & 258.40\\
\textbf{VAE-only} & 0.03 & 0.05 & 0.31 & 218.2\\
\textbf{No-residual} & 0.12 & 0.09 & 0.59 & 208.1\\
\textbf{Our Model} & \textbf{0.81} & \textbf{0.56} & \textbf{0.94} & \textbf{204.48}\\
\bottomrule

\end{tabular}
}
\caption{\label{results} Results for character F}
\end{table}

\begin{table}[h]
\centering
\scalebox{0.8}{
\begin{tabular}{llll|l}
\toprule
 & \textbf{V-measure}          & \textbf{Mutual Info}        & \textbf{F\&M} & \textbf{NLL}         \\
\midrule
\textbf{$\lambda$-only} & 0.60 & 0.62 & 0.73 & 268.40\\
\textbf{VAE-only} & 0.28 & 0.38 & 0.40 & 250.8\\
\textbf{No-residual} & \textbf{0.64} & \textbf{0.66} & \textbf{0.77} & 244.5\\
\textbf{Our Model} & 0.60 & 0.62 & 0.73 & 240.84\\
\bottomrule
\end{tabular}
}
\caption{\label{results} Results for character G}
\vspace{-0.2in}
\end{table}
\hspace{-0.2cm}\begin{table}[h]
\centering
\scalebox{0.8}{
\begin{tabular}{llll|l}
\toprule
 & \textbf{V-measure}          & \textbf{Mutual Info}        & \textbf{F\&M} & \textbf{NLL}         \\
\midrule
\textbf{$\lambda$-only} & 0.72 & 0.71 & 0.79 & 313.75\\
\textbf{VAE-only} & 0.32 & 0.32 & 0.40 & 254.2\\
\textbf{No-residual} & 0.90 & 0.97 & 0.94 & 258.8\\
\textbf{Our Model} & \textbf{0.92} & \textbf{1.01} & \textbf{0.96} & \textbf{249.81}\\
\bottomrule

\end{tabular}
}
\caption{\label{results} Results for character H}
\end{table}

\begin{table}[h]
\centering
\scalebox{0.8}{
\begin{tabular}{llll|l}
\toprule
 & \textbf{V-measure}          & \textbf{Mutual Info}        & \textbf{F\&M} & \textbf{NLL}         \\
\midrule
\textbf{$\lambda$-only} & 0.62 & 0.64 & 0.78 & 392.06\\
\textbf{VAE-only} & 0.29 & 0.38 & 0.40 & 323.5\\
\textbf{No-residual} & 0.70 & 0.83 & 0.74 & 329.25\\
\textbf{Our Model} & \textbf{0.75} & \textbf{0.84} & \textbf{0.87} & \textbf{323.04}\\
\bottomrule

\end{tabular}
}
\caption{\label{results} Results for character M}
\end{table}

\begin{table}[h]
\centering
\scalebox{0.8}{
\begin{tabular}{llll|l}
\toprule
 & \textbf{V-measure}          & \textbf{Mutual Info}        & \textbf{F\&M} & \textbf{NLL}         \\
\midrule
\textbf{$\lambda$-only} & 0.65 & 0.70 & 0.73 & 331.6\\
\textbf{VAE-only} & 0.30 & 0.45 & 0.40 & 265.2\\
\textbf{No-residual} & \textbf{0.74} & \textbf{0.81} & \textbf{0.82} & 270.11\\
\textbf{Our Model} & 0.69 & 0.75 & 0.75 & \textbf{264.23}\\
\bottomrule

\end{tabular}
}
\caption{\label{results} Results for character N}
\end{table}

\begin{table}[h]
\centering
\scalebox{0.8}{
\begin{tabular}{llll|l}
\toprule
 & \textbf{V-measure}          & \textbf{Mutual Info}        & \textbf{F\&M} & \textbf{NLL}         \\
\midrule
\textbf{$\lambda$-only} & 0.07 & 0.08 & 0.55 & 330.6\\
\textbf{VAE-only} & 0.03 & 0.04 & 0.34 & 247.1\\
\textbf{No-residual} & 0.06 & 0.07 & 0.53 & 251.32\\
\textbf{Our Model} & \textbf{0.46} & \textbf{0.32} & \textbf{0.78} & \textbf{246.02}\\
\bottomrule

\end{tabular}
}
\caption{\label{results} Results for character R}
\end{table}

\begin{table}[t]
\centering
\scalebox{0.8}{
\begin{tabular}{llll|l}
\toprule
 & \textbf{V-measure}          & \textbf{Mutual Info}        & \textbf{F\&M} & \textbf{NLL}         \\
\midrule
\textbf{$\lambda$-only} & 0.65 & 0.71 & 0.79 & 418.01\\
\textbf{VAE-only} & 0.31 & 0.45 & 0.42 & 364.2\\
\textbf{No-residual} & 0.72 & 0.78 & 0.82 & 369.5\\
\textbf{Our Model} & \textbf{0.72} & \textbf{0.79} & \textbf{0.84} & \textbf{364.21}\\
\bottomrule

\end{tabular}
}
\caption{\label{results} Results for character W}
\vspace{8.4in}
\end{table}

\end{document}